\documentclass[a4paper,conference]{IEEEtran}

\usepackage{cite}
\usepackage[pdftex]{graphicx}
\usepackage{alphabeta}
\usepackage{amsmath}
\usepackage{tabularx,booktabs}
\usepackage{amssymb}
\usepackage{calligra}
\usepackage{algorithmic}
\usepackage{array}

\usepackage{stfloats}
\usepackage{url}

\begin{document}

\title{Hierarchical Transformer for Survival Prediction Using Multimodality Whole Slide Images and Genomics}

% author names and affiliations use a multiple column layout for up to three different affiliations
% \author{
% \IEEEauthorblockN{Chunyuan Li}
% \IEEEauthorblockA{Department of Computer Science and Engineering\\
% University of Texas at Arlington\\
% Arlington, Texas 76013\\
% Email: chunyuan.li@mavs.uta.edu}
% \and
% \IEEEauthorblockN{Xinliang Zhu}
% \IEEEauthorblockA{Department of Computer Science and Engineering\\
% University of Texas at Arlington\\
% Arlington, Texas 76013\\
% Email: xinliang.zhu@mavs.uta.edu}
% \and
% \IEEEauthorblockN{Jiawen Yao}
% \IEEEauthorblockA{Department of Computer Science and Engineering\\
% University of Texas at Arlington\\
% Arlington, Texas 76013\\
% Email: jiawen.yao@mavs.uta.edu}
% \and
% \IEEEauthorblockN{Junzhou Huang}
% \IEEEauthorblockA{Department of Computer Science and Engineering\\
% University of Texas at Arlington\\
% Arlington, Texas 76013\\
% Email: jzhuang@exchange.uta.edu}
\author{\IEEEauthorblockN{Chunyuan Li,
Xinliang Zhu,
Jiawen Yao,
Junzhou Huang}
\IEEEauthorblockA{Department of Computer Science and Engineering\\
University of Texas at Arlington\\
Arlington, Texas 76013}
\IEEEauthorblockA{ Email: \{chunyuan.li, xinliang.zhu, 
jiawen.yao, jzhuang\}@uta.edu }
}

\maketitle

\begin{abstract}
Learning good representation of giga-pixel level whole slide pathology images (WSI) for downstream tasks is critical. Previous studies employ multiple instance learning (MIL) to represent WSIs as bags of sampled patches because, for most occasions, only slide-level labels are available, and only a tiny region of the WSI is disease-positive area. 
However, WSI representation learning still remains an open problem due to: 
(1) patch sampling on a higher resolution may be incapable of depicting microenvironment information such as the relative position between the tumor cells and surrounding tissues, while patches at lower resolution lose the fine-grained detail; 
(2) extracting patches from giant WSI results in large bag size, which tremendously increases the computational cost. To solve the problems, this paper proposes a hierarchical-based multimodal transformer framework that learns a hierarchical mapping between pathology images and corresponding genes. 
Precisely, we randomly extract instant-level patch features from WSIs with different magnification. Then a co-attention mapping between imaging and genomics is learned to uncover the pairwise interaction and reduce the space complexity of imaging features. Such early fusion makes it computationally feasible to use MIL Transformer for the survival prediction task. 
Our architecture requires fewer GPU resources compared with benchmark methods while maintaining better WSI representation ability.
We evaluate our approach on five cancer types from the Cancer Genome Atlas database and achieved an average c-index of $0.673$, outperforming the state-of-the-art multimodality methods. 
%Visualization of the ... further demonstrates ...
\end{abstract}

\IEEEpeerreviewmaketitle

\section{Introduction}

Survival analysis is an essential task in cancer study and diagnosis. Instead of observing the biopsies under the microscope, whole slide pathology images (WSI) are widely used to present tumor growth and morphology information. Automatically analyzing histology provides valuable help for histologists on precision medicine and reduces diagnosis bias. However, WSIs are typically in multi-gigapixel level with high morphological variance, which brings a tremendous challenge to the feature representation of WSIs.

% WSI representation: not end-to-end, not global optimical. MIL. fixed magnification: loose morphology information at larger view.
Many solutions have been proposed to solve the representation problem of WSIs \cite{yao2016imaging,zhu2016lung}. A widely used strategy is patch-based processing, where thousands of patches are extracted from a WSI using either a non-overlap sliding window or a random sample method \cite{zhu2017wsisa,zhu2016deep,yao2017deep}. In the clinics, the tumor area usually only takes a tiny portion of the WSI, which means most patches sampled from the gigapixel WSI are unrelated to survival prediction. Unfortunately, the patch-wise annotation for WSIs is laborious and even infeasible in most situations. Many researchers consider weakly supervised learning approaches to solve the problem with a lack of annotations. Zhu et al. \cite{zhu2017wsisa} proposed a two-stage framework for survival prediction on WSIs without annotations. They applied K-means clustering to group the patches according to visual appearance and then aggregated patches from the selected clusters for survival prediction. Recent approaches \cite{ilse2018attention,yao2020whole,chen2021multimodal,li2021dual} use multiple instance learning (MIL) \cite{maron1998framework} to formulate the WSI representation, where each patient is considered as a \textit{bag} containing a set of \textit{instances} of patches. A \textit{bag} is annotated as disease-positive if there is any disease-positive patch in the \textit{bag}. Patch features are integrated with fully connected (FC) layers for specific tasks. On the basis of attention-based MIL, Chen \textit{et al.}, \cite{chen2021multimodal} first proposed a Multimodal Co-Attention Transformer (MCAT) architecture that investigates early fusion mechanisms for identifying informative patches. MCAT learns the long-range relationship between image and gene using attention mechanism and visually shows the interpretability of multimodal interaction. However, MCAT uses non-overlapping patches sampling and maintains a bag size of over ten thousand, which essentially increases the computation requirement during training and is time-consuming.  %加点ref 再写点?

% option: 写点multi modality methods and limitations?

\begin{figure}[!t]
    \centering
    \includegraphics[width=2.5in]{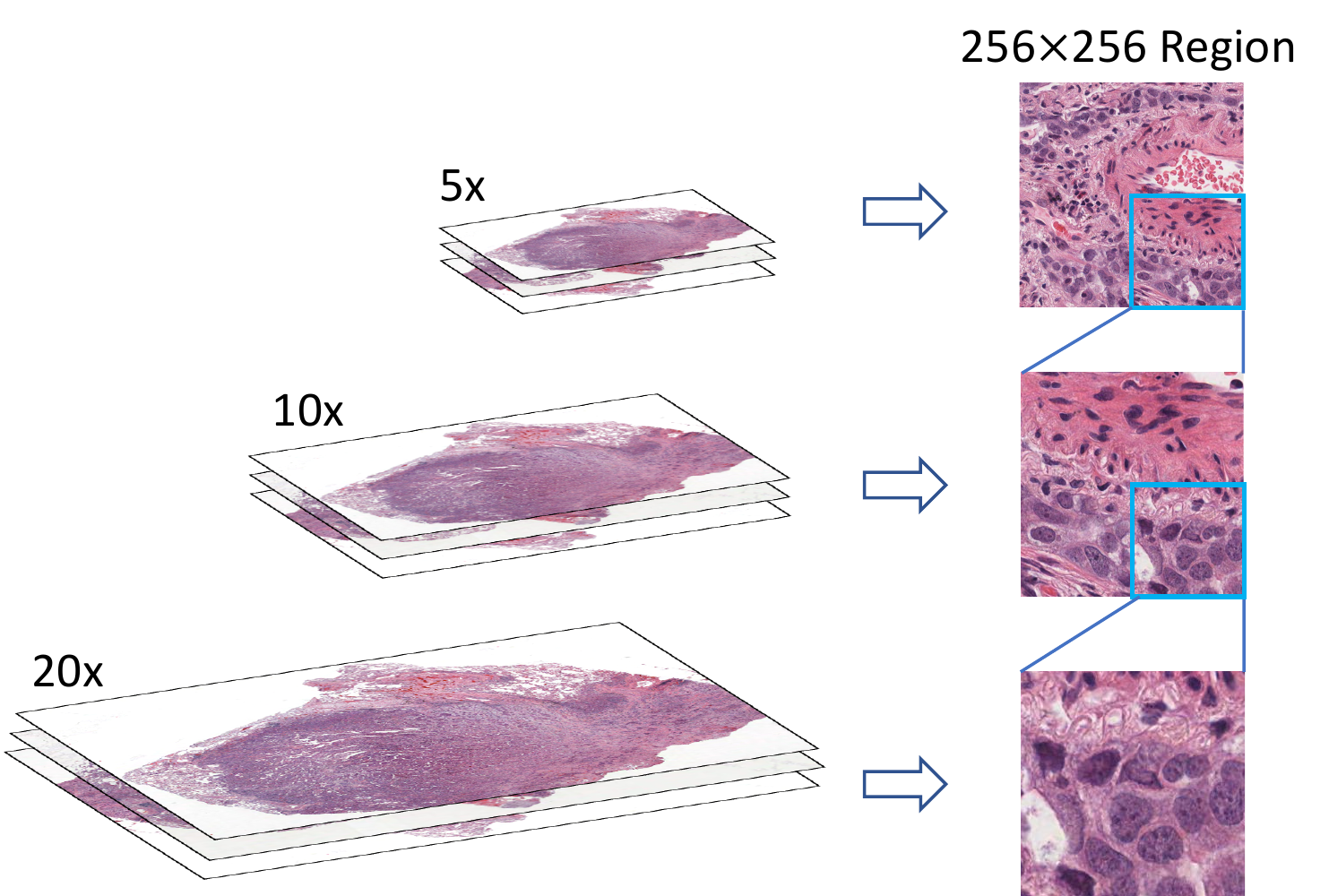}
    \caption{Hierarchy structure and example patches. Left: WSIs at $5x$, $10x$, and $20x$ magnification. Right: Patches extracted from WSIs at different magnifications. The blue box indicated the corresponding region in upper magnification.}
    \label{fig:hierarchy_structure}
\end{figure}

% limitation: region size and influence.
Two major limitations exist in the above approaches for WSI representation. Firstly, current studies use a fixed scale for patch extraction. The widely used patch sizes, either $[256\times 256]$ or $[512\times 512]$, have limited ability to capture coarser level tissue morphology characteristics, such as tumor shape, size, and circularity that are essential to determine grades of glioma \cite{wang2019machine,beck2011systematic,kothari2012biological,chan2019histosegnet}. As shown in Figure \ref{fig:hierarchy_structure}, WSIs are scanned at $20\times$ with a resolution of $0.5\mu p$ per pixel. The $256\times 256$ patches sampled from WSI at $20\times$ magnification can capture a small set of cell characteristics and interactions, while patches extracted from downsampled WSIs at $10\times$ or $5\times$ present rich microenvironmental information between different sets of cells or tissue. The multi-scale patches depict significant inter- and intra-instance differences in shape, size, and pattern. Thus they better represent the overall heterogeneity of the tumor microenvironment \cite{graham2019hover}. %加点ref
Secondly, existing approaches use VGG and Resnet models pre-trained on ImageNet to obtain patch features. Compared with vision Transformer (ViT), CNN models do not incorporate spatial location information and maintain less global information. The different structures result in quantitative variability in out-of-distribution generalization ability \cite{raghu2021vision,zhang2021delving}. %再多写点
% transformer encoder on gene. optional-提一下transformer encoder处理基因

To address these challenges, we present a weakly supervised, Transformer-based hierarchical architecture named Hierarchical MIL Transformer (HiMT) that learns a mapping between genetics and WSIs at different scales for survival prediction. Specifically, HiMT randomly samples a few patches from different scales, then constructs bag representation using processed images and gene categories. HiMT learns a co-attention imaging-genomic mapping to reduce the space complexity of patches feature. The gene-guided visual concept can be used on various tasks with proper FC layers. Here we demonstrate the power with the survival prediction task. 
During the patch sampling stage, our hierarchy strategy ensures the patient bag has a reasonable size of approximately 3000, while our benchmark architecture \cite{chen2021multimodal} collected an average bag size of 15,000, which requires a plenty of GPU resources during training. 
Our main contributions can be summarized as follows. 
\begin{itemize}
    \item We introduce a hierarchical mechanism to identify the informative instance across different scales using genetic features as queries.  
    \item We use a pre-trained ViT for feature extraction and showed in ablation studies that even with a single WSI scale, features extracted from $80\%$ fewer patches than MCAT could achieve competitive performance. 
    \item Instead of using a non-overlapping patch sampling procedure, we randomly sampled fewer patches and essentially reduced the computation stress while maintaining outperforming results. 
\end{itemize}
We evaluate HiMT on five popular TCGA cancer datasets. Our results outperform the state-of-the-art MIL models on imaging-genetic data mapping tasks, which show that our architecture can capture representative features with vastly reduced feature space. 
Experiments further show that although the number of patches dramatically reduced compared to previous work \cite{chen2021multimodal} which collects on average over ten thousand patches from each WSI, our architecture performs better in survival prediction. 
In addition, we conduct an ablation study on the five TCGA datasets and demonstrate that HiMT improves with patches from different magnification. 
Our code is publicly available at \url{https://github.com/chunyuan1/HiMT}. 

\section{Related Work}

% \textbf{Feature representation of WSI.}

\subsection{Weakly Supervised Learning for WSI Analysis} 
Recent works have developed many weakly supervised models for medical image diagnosis to overcome the unavailability of manual annotation in clinical data \cite{hou2016patch,yao2020whole,lu2021ai}. Edwards and Storkey and Zaheer et al. first proposed network architecture on set-based data \cite{edwards2016towards, zaheer2017deep}. Ilse et al. extend the set-based concept by employing an attention mechanism and applying it on WSIs \cite{ilse2018attention}. Yao et al. incorporated attention-based MIL on clustered phenotype and achieved promising results \cite{yao2020whole}. 
Instead of only focusing on the instance-level feature, Chen et al. proposed co-attention MIL Transformer that learns an interpretable mapping and visually demonstrates the relationship between imaging and genetics \cite{chen2021multimodal}.
However, the approaches at a single scale cannot maintain high-level morphological features. We believe that fewer patches with the multi-scale patches feature can achieve promising results compared to tens of thousands of patches.
% A deep fully connected network (DeepSurv) \cite{katzman2016deep} was first proposed to represent the nonlinear risk function, which outperform the traditional Cox hazard model. Zhu et al. \cite{zhu2016deep} applied deep survival model on pathological images and achieve great improvement. Yao et al. \cite{yao2017deep} further intrgrated the genome modality with DeepConvSurv to predict survival on multi-modality data. The above methods are based on pre-selected regions of interest (ROI) patches, which reduce the size of image tiles and save the need of computational power. However, the selected patches set might not properly reflect tumor morphology. 

% \textbf{Multiple Instance Learning for WSIs Analysis.} 

\subsection{Self-Attention and Vision Transformer} 
Solving computer vision tasks using non-convolutional neural networks has been an active research area \cite{vaswani2017attention,dosovitskiy2020image,touvron2021training}. The original ViT takes $256\times 256$ images as input and processes it as a sequence of $[16\times 16]$ tokens with positional information. Since ViT was first proposed by Dosovitskiy et al., plenty of recent work is developed to analyze aspects of ViT, such as robustness \cite{bhojanapalli2021understanding}, effects of self-attention \cite{caron2021emerging}, designing improved ViT models \cite{liu2021swin, chen2021crossvit} and comparison with CNN \cite{raghu2021vision,zhang2021delving}. ViT has been proved for its out-of-distribution generalization and outstanding feature representation due to the accessibility to global information at an early stage \cite{raghu2021vision,zhang2021delving}. In the research of medical images, pre-trained Resnet is widely used for feature extraction, while ViT is rarely explored. The reason may be that training a transformer model from scratch requires an immense amount of data and intensive computational resources. We highlight that the out-of-distribution generalization ability of ViT makes the ImageNet-pretrained ViT model been able to extract high-quality representative features from the medical image. Based on the fact that ViT model relationships between the tokens, we design a hierarchy model that allows ViT to learn interaction at different scales of the WSIs.

% 一点Transformer 似乎用不到
% outstanding performance on describing correlation between tokens in a sequence and modelling long distance information. Self-attention mechanism is adopted in transformer to capture the pairwise relationship between input tokens. However, traditional Transformer model are limited by memory intensity and its quadratic time complexity. It might be incapable to handle long sequences. Therefore, it is not suitable for large size images such as WSIs or WSI patches. 

\section{Method}
\label{sec:method}

\begin{figure*}[htb]
\centering
  \includegraphics[width=16cm]{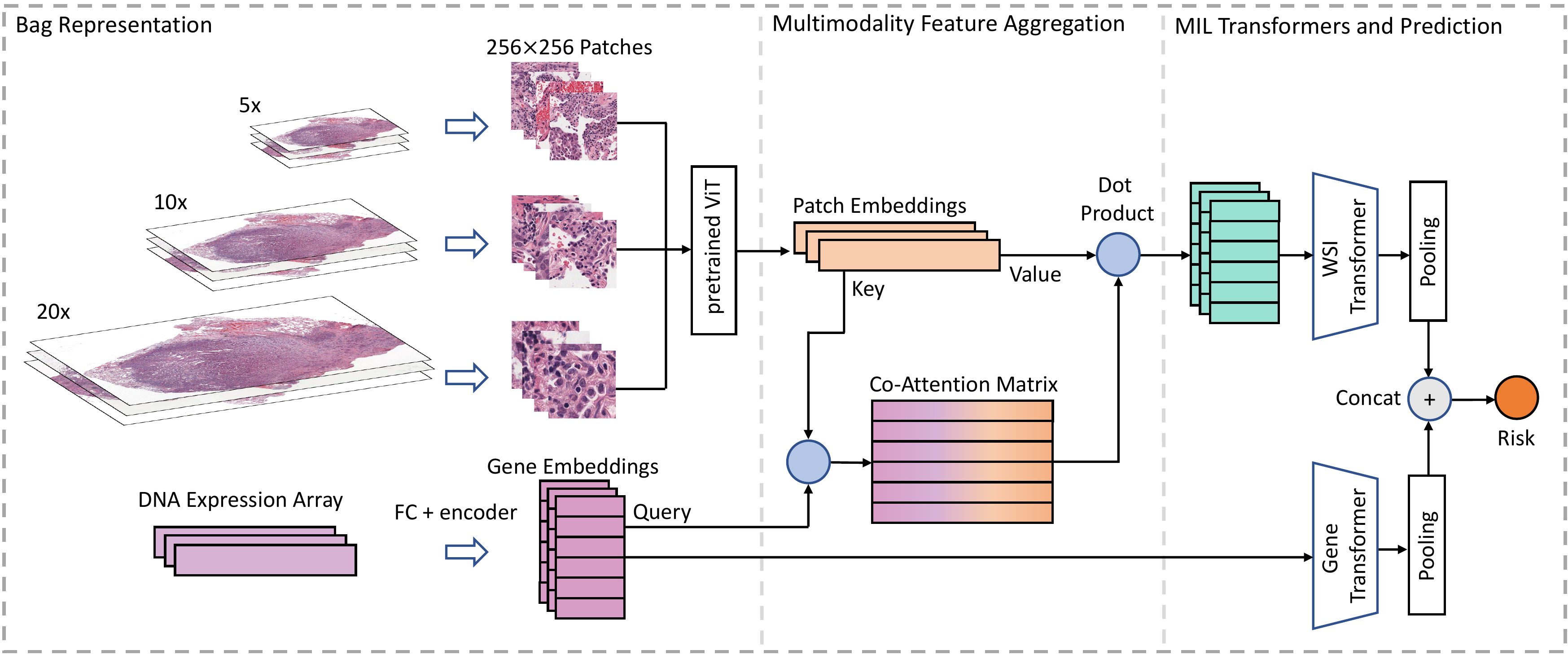}
%   WSI Transformer 已改 
  \caption{Framework overview of HiMT contains three stages. Bag representation: 1) Patches sampled from WSIs with different magnification are sent into a pre-trained ViT model to extract instance-level patch feature; 2) DNA microarray are categorized according to functional expression. Early feature fusion: This stage learns a gene-guided co-attention mapping that directly models the pairwise relationship between pathological images and genomics. The GCA is used to reduce the sequence length of WSI bags, leading to more flexibility in choosing aggregation strategies in the following stage. MIL transformer and survival prediction: applies set-based MIL transformer on gene-guided imaging feature and genomics for survival prediction. }
  \label{fig:framework}
\end{figure*}

Figure \ref{fig:framework} depicts an overview of HiMT architecture. 
Given a set of $\Omega$ patients ${X_i}, i=1, ..., \Omega$, each patient has its feature vector $(Surv_i, \delta_i)$. The $Surv_i$ indicates the time until the event of interest occurs, and vital status $\delta_i\in \{0,1\}$ indicates whether the event is censored or uncensored (death occurs). The objective is to predict a corresponding target variable $o_i$ for patient $X_i$ from the imaging data.

\subsection{Patient Bag Construction}
This stage aims to formulate the problem by constructing bags using pathology images and genomics. Multiple Instance Learning (MIL) uses bags as the data samples, each containing a set of unordered instances. In our work, patient $X_i$ is a bag of instance, which contains feature vectors $\{x_1,\ldots ,x_M\}\in \mathbb{R}^{M\times d}$ from all WSIs of $X_i$ and functional gene sets $\{G\}_i$, where the number of instances $M$ various for each image bag.
% $\{W_{ij}\}_{j=1}^{K_i}$ 

\textbf{Hierarchy Feature Embedding using ViT.} % todo: 多写点 详细
Due to the considerable size, it is infeasible to use WSIs as input. We extract image features by sampling patches at different magnifications from all WSIs belonging to the same patients. Instead of sampling non-overlapping patches, we randomly generate patches from multi-scale WSIs. A thousand patches are extracted randomly from WSIs at $5\times$, $10\times$, and $20\times$ magnification, respectively, with a fixed size of $256\times 256\times 3$. The patches that contain background are excluded. The above procedure results in three thousand patches for each WSI. The number of patches is about eight times less than the non-overlapping sampling strategy, while it contains more morphology information at different resolutions. Then a ViT-L16 model \cite{dosovitskiy2020image} (pre-trained on ImageNet) is used to extracts $d_k$-dim feature embeddings $h\in\mathbb{R}^{d_k\times 1}$. The $M$ extracted patch embeddings from all WSIs of patient $X_i$ are packed into a bag $H_i$, where $H_{bag}\in \mathbb{R}^{M\times d_k}$ and $M$ is approximately $3000$. 

\textbf{Genomic Feature Embedding.} 
The heterogeneity within a gene set limited the utility of the gene database. A Gene alone from the gene set cannot coherently describe the biological impact. To use the semantic information within the gene set, genes that contribute to the same biological function are categorized together and end up with $N$ gene sets. Let $D_i=\{d_{j}\}$ be the gene expression array of patient $X_i$ and $\{S_n\}^N_{n=1}$ represents the functional categories according to \cite{liberzon2015molecular, subramanian2005gene}. For each gene expression in $\{d_{j}\}$, if its attribute $attr_{j}\in S_n$, $d_{j}$ is assigned to gene set $g_n$. In the resulting gene sets $\{g_n\}_{n=1}^N$, each set $g_n$ has various size. We then apply a FC layer over all the gene sets to obtain the final genomic embeddings $\{G_n\in \mathbb{R}^{d_k\times 1}\}^N_{n=1}$ and pack it into a bag $G_{bag}\in \mathbb{R}^{N\times d_k}$. In our experiments, $N$ is set to $6$, indicating the following functional categories: Tumor Suppression, Oncogenesis, Protein Kinases, Cellular Differentiation, Transcription, and Cytokines and Growth.

\subsection{Multimodality Feature Aggregation}
To capture interpretable interactions between genes and pathological images, we use an early fusion mechanism, Genomic-Guided Co-Attention (GCA) \cite{chen2021multimodal}, to discover the genotype-phenotype interactions from the tumor microenvironment.  Inspired by Transformer attention \cite{vaswani2017attention}, GCA directly model the pairwise relationship by learning a co-attention matrix $A_{coa}$ between patch embeddings in $H_{bag}\in \mathbb{R}^{M\times d_k}$ and gene embeddings $G_{bag}\in \mathbb{R}^{N\times d_k}$. Then $A_{coa}$ is used to map the $H_{bag}$ to a set of gene-guided visual concept $\widehat{H}_{bag}$. The mapping can be expressed as

\begin{equation}
    \begin{aligned}
        CoA(G,H)=softmax(\frac{QK^\top}{\sqrt{d_k}})\\
        =softmax(\frac{W_qGH^\top W_k^\top}{\sqrt{d_k}})W_vH\\
        \to A_{coa}W_vH \to \widehat{H}
    \end{aligned}
\end{equation}
where weight matrices $W_q, W_k, W_v\in \mathbb{R}^{d_k\times d_k}$ are trainable parameters to map the query $G_{bag}$ and key-value pairs $(H_{bag},H_{bag})$ to an output. In the experiment, bag size of $Q$ is much smaller than that of $K$ and $V$. Consequently, GCA can largely reduce the complexity of WSI bags.

\subsection{MIL Transformers with Survival Prediction}
With image embeddings $\widehat{H}_{coa}\in \mathbb{R}^{N\times d_v}$ and genomic embeddings $G_{bag}\in \mathbb{R}^{N\times d_k}$ as inputs, we use two set-based MIL Transformers $\mathcal{T}_H$ and $\mathcal{T}_G$ to aggregate image and genomic features, respectively \cite{zaheer2017deep,vaswani2017attention,lee2019set,chen2021multimodal}. This step can be written as

\begin{equation}
    \begin{split}
    &\mathcal{E}^{(l)}(H^{(l)})=\zeta^{(l)}(\psi^{(l)}(\{ \phi^{(l)}(x_i):h_i^{(l)}\in H^{(l)} \})) \\ 
    &\mathcal{F}^{(L)}(H^{(L)})=\zeta^{(L)}(\rho^{(L)}(\{ \phi^{(L)}(x_i):h_i^{(L)}\in H^{(L)} \})) \\ 
    &\mathcal{T}(X)=\mathcal{F}^{(L)}(\mathcal{E}^{(L-1)}(...\mathcal{E}^{(1)}\{(x_i):x_i\in X\}))
    \end{split}
\end{equation}

Among the above equations: $h_i^{(l)}$ is an arbitrary instance in the set $H^{(l)}$ at hidden layer $l$. $\phi:\mathbb{R}^{d_{in}}\to \mathbb{R}^{d_{out}}$ is an instance-level functions. $\psi^{(l)}$ is the self-attention layer. $\rho:\mathbb{R}^{m\times d_{out}}\to \mathbb{R}^{d_{out}}$ is a permutation-invariant function that aggregate instances to bag-level feature. $\zeta:\mathbb{R}^{d_{out}}\to \mathbb{R}^{\text{\# class}}$ is a bag-level classifier that undertake target-specific task. In our work, $\zeta$ is a position-wise FC layer and is used to estimate risk score for survival analysis. $\mathcal{E}^{(l)}$ is an encoder block. $\mathcal{F}^{(L)}$ is the MIL network that perform global pooling at the last layer $L$. Precisely, $\psi^{(l)}$ can be written as the set function which is permutation-equivariant:

\begin{equation}
    \psi^{(l)}(\{h_i^{(l)}\}^M_{i=1})=\{\sum^M_{i=1} \frac{\exp{(h_i^{(l)}h_j^{(l)\top})}}{d_k\sum_j\exp{(h_i^{(l)}h_j^{(l)\top})}}\cdot h_i^{(l)}\to h_i^{(l+1)} \}
\end{equation}

The expression indicates that Transformer is a generalization of shallow set-based data structure. $\rho_H$ and $\rho_G$ are implemented following \cite{ilse2018attention}:

\begin{equation}
\begin{split}
    & \phi^{(L)}(h_i^{(L)})=W_\phi h_i^{(l)} \\
    & \rho^{(L)}(\{h_i^{(L)}\}^M_{i=1}) =  \sum^M_{i=1}a_i\phi^{(L)}(h_i^{(L)})\to h^{(l)} \text{ where}\\
    & a_i = \frac{\exp\{ W_\rho(\tanh (V_\rho h_i^{(L)\top})\odot \text{sigm}(U_\rho h_i^{(L)\top})) \} }
    { \sum^M_{j=1}\exp\{ W_\rho (\tanh (V_\rho h_j^{(L)\top})\odot \text{sigm} (U_\rho h_j^{(L)\top}) ) \} } \\
    & \zeta^{(L)}(h^{(L)}) = W_\zeta h^{(L)}
\end{split}
\end{equation}

in which $W_\phi$, $W_\rho$, $V_\rho$, $U_\rho$, $W_\zeta \in \mathbb{R}^{d_v\times d_v}$ are trainable parameters, $\phi^{(L)}$ is a FC layer processing instance-level feature. $\rho^{(L)}$ is a bag-level attention pooling layer and $a_i$ controls weight of embedding $h_i^{(L)}$ contributing to bag-level $h^{(L)}$. 

Finally, HiMT integrates output of $\mathcal{T}_H$ and $\mathcal{T}_G$ to bag-level features by concatenating, denoted as $[\zeta_h^{(L)}(h^{(L)}), \zeta_g^{(L)}(g^{(L)})]$. Several FC layers are added in the end to obtain the final risk score $o_{risk}$.

\subsection{Loss Function}

For $i$-th patient, the output hazard risk is denoted as $o_{risk,i}$. Let continuous random variable $T$ represent overall survival time. Following the discovery from chen et al. \cite{chen2021multimodal} that loss function is mini-batch dependent, we model the survival prediction problem using discrete time intervals. Given right-censored survival data, we partition the time scale into non-overlapping intervals: $[t_0,t_1),[t_1,t_2),[t_2,t_3),[t_3,t_4)$ depending on the quantiles of survival time. For each patient, the discrete event time with the continuous event time $T_{i,cont}$ can be defined as
\begin{equation}
    T_i=r\text{ if }T_{i,cont}\in [t_r,t_{r+1}) \text{ for }r\in \{0,1,2,3\}
\end{equation}
Denote the discrete time ground truth of $i^{th}$ patient as $Y_i$. For a patient with bag-level risk $o_{risk,i}$, the hazard function that measures the probability of patient die at time $r$ can be expressed as
\begin{equation}
    f_{hazard}(r|o_{risk,i})=P(T_i=r|T_i\geq r,o_{risk,i})
\end{equation}
while the survival function that estimate the probability of a patient live longer than time point $R$ can then be defined as
\begin{equation}
    \begin{split}
        f_{surv}(r|o_{risk,i})=P(T_i > r|o_{risk,i})\\
        =\prod^r_{u=1}(1-f_{hazard}(u|o_{risk,i}))
    \end{split}
\end{equation}
When updating the parameters of discrete survival model, we use log likelihood \cite{zadeh2020bias} that consider the vital status of patients:
\begin{equation}
    \begin{split}
        L=-c_i\cdot \log (f_{surv}(Y_i|o_{risk,i}))\\
        -(1-c_i)\cdot \log (f_{surv}(Y_i-1|o_{risk,i}))\\
        -(1-c_i)\cdot \log (f_{surv}(Y_i|o_{risk,i}))\\
    \end{split}
\end{equation}
Weighted sum is applied on $L$ and $L_{uncensored}$ that up-weight the contribution of uncensored patients.
\begin{equation}
    L_{surv}=(1-\beta)\cdot L+\beta \cdot L_{uncensored}
\end{equation}
where the uncensored loss is computed as
\begin{equation}
    \begin{split}
        L_{uncensored}=-(1-c_i)\cdot \log(f_{surv}(Y_i-1|o_{risk,i}))\\
        -(1-c_i)\cdot \log(f_{hazard}(Y_i|o_{risk,i}))
    \end{split}
\end{equation}

\section{Experiments}
\label{sec:exp}

\subsection{Dataset Description}

To show the performance of our architecture, we conduct experiments on five largest cancer datasets from The Cancer Genome Atlas (TCGA): Bladder Urothelial Carcinoma (BLCA), Breast Invasive Carcinoma (BRCA), Glioblastoma Multiforme (GBM) $\&$ Brain Lower Grade Glioma (LGG), Lung Adenocarcinoma (LUAD), Uterine Corpus Endometrial Carcinoma (UCEC). The numbers of WSIs and patients in each dataset are shown in Table \ref{tab:dataset}. For each WSI, $1000$ patches are randomly sampled form each hierarchy layer. 5-fold cross-validation is performed on each dataset.

\begin{table}[ht]
 \caption{The number of WSIs and patients in each TCGA dataset.}
  \centering
  \begin{tabular}{c|c c c c c}
%   \hline
    \toprule
    Dataset & BLCA & BRCA & GBMLGG & LUAD & UCEC\\
    \hline
    \# patients & 373 & 957 & 569 & 453 & 480 \\
    \# WSIs & 437 & 1022 & 1011 & 515 & 538 \\
    \bottomrule
  \end{tabular}
  \label{tab:dataset}
\end{table}

\begin{table*}[ht]
 \caption{Performance comparison with different methods using C-index values.}
  \centering
  \begin{tabular}{c|c c c c c c}
%   \hline
    \toprule
    Methods     & BLCA & BRCA & GBMLGG & LUAD & UCEC & Overall\\
    \hline
    DeepMISL (WSI Only) \cite{ilse2018attention} &0.536$\pm$0.038 & 0.564$\pm$0.050 & 0.787$\pm$0.028 & 0.559$\pm$0..060 & 0.625$\pm$0.057 & 0.614\\
    DeepMISL (Concat)  & 0.605$\pm$0.045 &0.551$\pm$0.077 & 0.816$\pm$0.011 & 0.563$\pm$0.050 & 0.614$\pm$0.052 & 0.630\\
    DeepMISL (Bilinear Pooling) & 0.567$\pm$0.034 & 0.536$\pm$0.074 & 0.812$\pm$0.005 & 0.578$\pm$0.036 & 0.562$\pm$0.058 & 0.611\\
    
    DeepAttnMISL (WSI Only) \cite{yao2020whole} & 0.504$\pm$0.042 & 0.524$\pm$0.043 & 0.734$\pm$0.029 & 0.548$\pm$0.050 & 0.597$\pm$0.059 & 0.581\\
    DeepAttnMISL (Concat) & 0.611$\pm$0.049 & 0.545$\pm$0.071 & 0.805$\pm$0.014 & 0.595$\pm$0.061 & 0.615$\pm$0.020 & 0.634\\
    DeepAttnMISL (Bilinear Pooling) & 0.575$\pm$0.032 & 0.577$\pm$0.063 & 0.813$\pm$0.022 & 0.551$\pm$0.038 & 0.586$\pm$0.036 & 0.621\\
    
    MCAT \cite{chen2021multimodal} & 0.624$\pm$0.034 & 0.580$\pm$0.069 & 0.817$\pm$0.021 & \textbf{0.620$\pm$0.032} & 0.622$\pm$0.019 & 0.653\\
    % MCAT-reproduce & 0.628$\pm$0.029 & 0.603$\pm$0.069 & 0.811$\pm$0.032 & 0.609$\pm$0.047 & 0.554$\pm$0.057 \\
    \hline
    ours & \textbf{0.660$\pm$0.021} & \textbf{0.606$\pm$0.028}& \textbf{0.823$\pm$0.019} & 0.616$\pm$0.016 & \textbf{0.658$\pm$0.047} & \textbf{0.673}\\
    \bottomrule
  \end{tabular}
  \label{tab:cindex}
\end{table*}

\subsection{Implementation Details}
HiMT is trained with one Geforce GTX 1080 Ti GPU. The functional signatures to category the gene embeddings are obtained from \cite{liberzon2015molecular}. After collecting the six gene categories, several FC layers are applied to convert the gene categories into similar lengths, followed by one encoder layer. Patch features are extracted by ViT-Large model pre-trained on ImageNet-21k with $16\times 16$ token embedding strategy \cite{dosovitskiy2020image}, followed by a FC layer that obtains a feature vector of length $1024$ for each patch. The third phase of HiMT uses two Vision Transformer from \cite{dosovitskiy2020image} with an attention pooling layer to deal with WSI and genomic features, respectively. The processed logits are then concatenated together and go through several FC layer with sigmoid function to obtain the final risk score $o_{risk}$.
During training, we follow the experimental settings in \cite{chen2021multimodal} that use Adam optimizer \cite{kingma2014adam} with a learning rate of $2\times 10^{-4}$ and weight decay of $1\times 10^{-5}$. Batch size is set to $1$ due to the various bag size. 

\subsection{Evaluation metrics}
Concordance index (C-index) is a popular metric in evaluating survival prediction models~\cite{zhu2016deep,zhu2017wsisa,yao2020whole}. We also use it as evaluation metric in our experiments. The C-index quantifies the correlation of risk score and survival time, calculated as follows:
\begin{equation}
    c=\frac{1}{n}\sum_{i\in \{1...N|\sigma_i=1\}}\sum_{s_j > s_i} I[X_i\hat{\beta}>X_j\hat{\beta}]
\end{equation}
where $n$ is the number of comparable pairs, $I[.]$ is the indicator function and $s.$ is the actual observation. The value of C-index ranges from 0 to 1. The larger CI value means the better prediction performance of the model and vice versa. 

\subsection{Experimental Results}
We compare our architecture with several state-of-the-art methods for survival prediction using the same 5-fold cross-validation splits, training hyper-parameters, and loss function. 
\begin{itemize}
    \item DeepMISL \cite{ilse2018attention}: Deep multiple survival learning is a set-based network that first applies global attention into MIL algorithm.
    \item DeepAttnMISL \cite{yao2020whole}: DeepAttnMISL first applies K-Mean clustering to group instant-level feature into phenotypes. Then an attention-based global pooling is used to aggregate the risk score of each patient.
    \item MCAT \cite{chen2021multimodal}: A state-of-the-art multimodal survival architecture that incorporate self-attention mechanism to learn an interpretable mapping between imaging and genomics in early fusion manner.
\end{itemize}

Table \ref{tab:cindex} presents the results using the above methods on five cancer datasets. 
Compared with WSI-based MIL methods (DeepMISL, DeepAttnMISL), HiMT increases the overall c-index by $9.6\%$ and $15.8\%$. 
Against the similar methods with multimodal approaches, HiMT achieves $6.8\%$, $6.2\%$, and $3.1\%$ advancement over DeepMISL, DeepAttnMISL, and MCAT, respectively. DeepMISL and DeepAttnMISL both attain more reasonable results than baseline models that only use WSIs. However, DeepMISL shows a considerable variance among all experiments with $p>0.1$, indicating the result is unreliable. DeepMISL exhibits a marginal performance boost over the baseline method with statistically significant for the BLCA ($p<0.01$). Unlike the late fusion strategy used in DeepMISL and DeepAttnMISL, MCAT proposed an early fusion mechanism called co-attention and performs notable progress over previous work with $p<0.01$ for BLCA, GBMLGG, and UCEC. Our method achieves the best average c-index among all the approaches with $p<0.01$ for all five cancer datasets.
Specifically, HiMT achieves the highest c-index in 4 out of 5 cancer datasets in comparison concerning each cancer. This indicates that the hierarchy architecture in HiMT can capture representative features from medical images while also requiring less computational resources. 

\begin{figure}[!t]
    \centering
    \includegraphics[width=2.8in]{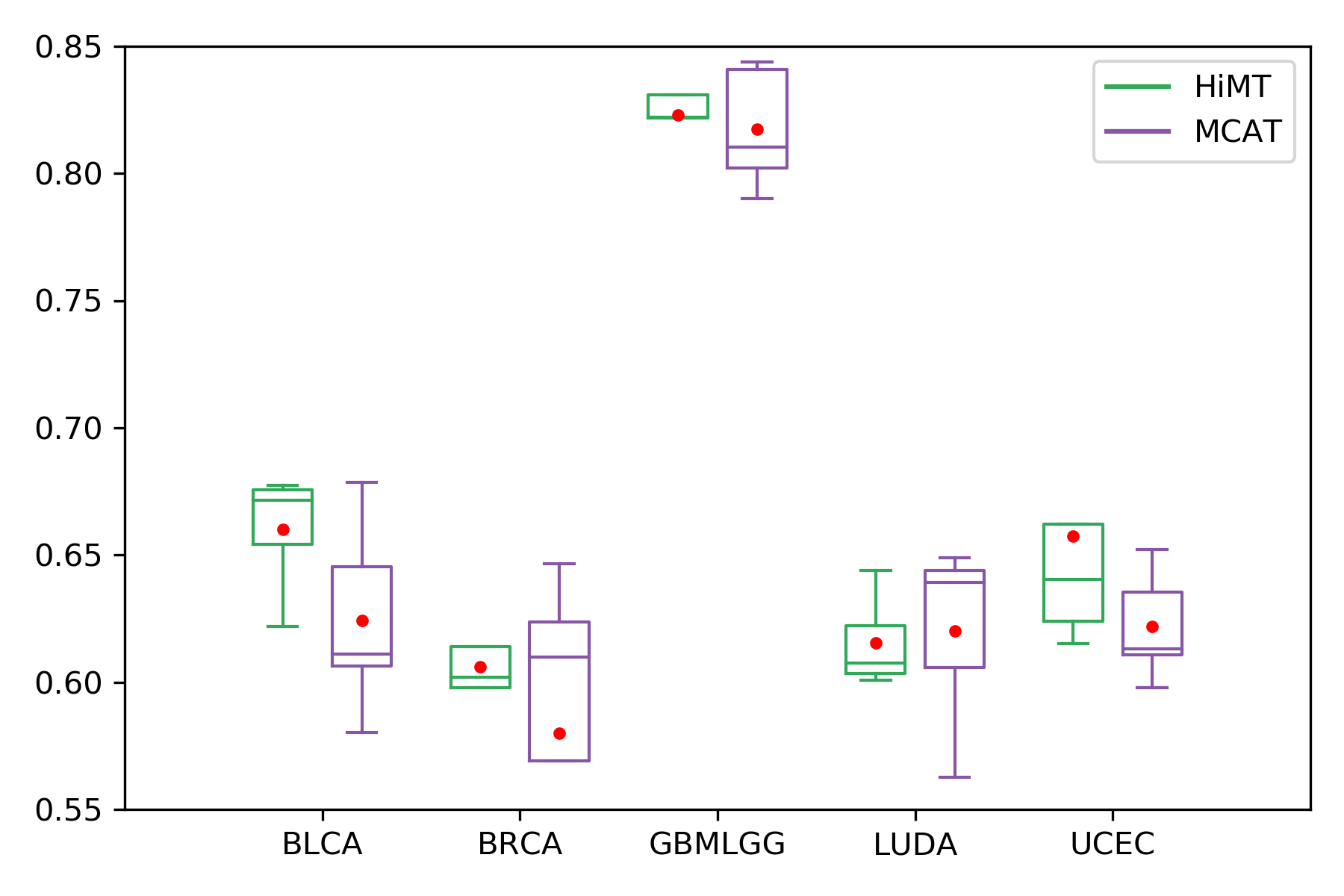}
    \caption{Boxplots of C-index values of five cancer datasets compared with MCAT. The red dot indicates the mean C-index of the 5-fold cross-validation.}
    \label{fig:boxplot}
\end{figure}

The boxplot of C-index in Figure \ref{fig:boxplot} depicts the variation of C-index values among the 5-fold and further demonstrate that our algorithm not only performs better in terms of mean C-index but also is more robust than previous work. 

\begin{table*}[!t]
 \caption{Ablation study assessing C-index performance with respect to different magnification settings. 1000 $256\times 256$ patches are sampled from each magnification for experiments shown below.}
  \centering
  \begin{tabular}{c|c c c c c c}
    \toprule
    Methods     & BLCA & BRCA & GBMLGG & LUAD & UCEC & Overall (C-index) \\
    \hline
    Patches at $5\times$  & 0.636$\pm$0.022 & \textbf{0.612$\pm$0.050} &0.806$\pm$0.006  & 0.593$\pm$0.061& 0.556$\pm$0.045 & 0.641 \\
    
    Patches at $10\times$  & 0.634$\pm$0.010 & 0.586$\pm$0.047 &0.804$\pm$0.009  & 0.598$\pm$0.045& 0.635$\pm$0.071 & 0.651 \\
    
    Patches at $20\times$  & 0.638$\pm$0.028 & 0.609$\pm$0.046 & 0.821$\pm$0.027 & 0.609$\pm$0.030 & 0.638$\pm$0.031 &  0.663 \\ %0.660
    
    Patches at $20\times$ and $10\times$ & 0.626$\pm$0.018 & 0.592$\pm$0.029 & 0.805$\pm$0.008 & \textbf{0.617$\pm$0.048} & 0.645$\pm$0.078 & 0.657 \\
    
    Patches at $20\times$, $10\times$ and $5\times$ & \textbf{0.660$\pm$0.021} & 0.606$\pm$0.028& \textbf{0.823$\pm$0.019} & 0.616$\pm$0.016 & \textbf{0.658$\pm$0.047} & \textbf{0.673} \\
    \bottomrule
  \end{tabular}
  \label{tab:ablation-c}
\end{table*}

\begin{table*}[!t]
 \caption{Ablation study assessing AUC values with respect to different magnification settings. 1000 $256\times 256$ patches are sampled from each magnification for experiments shown below.}
  \centering
  \begin{tabular}{c|c c c c c c}
    \toprule
    Methods     & BLCA & BRCA & GBMLGG & LUAD & UCEC & Overall (AUC) \\
    \hline
    Patches at $5\times$  & 0.664$\pm$0.051 & \textbf{0.648$\pm$0.055} &0.843$\pm$0.016  & 0.612$\pm$0.073& 0.583$\pm$0.047 & 0.670 \\
    
    Patches at $10\times$  & 0.659$\pm$0.031 & 0.611$\pm$0.045 &0.844$\pm$0.016  & 0.616$\pm$0.050& 0.651$\pm$0.082 & 0.676 \\
    
    Patches at $20\times$  & 0.663$\pm$0.055 & 0.626$\pm$0.040 & \textbf{0.862$\pm$0.028} & 0.624$\pm$0.047 & 0.655$\pm$0.028 &  0.686 \\ %0.660
    
    Patches at $20\times$ and $10\times$ & 0.654$\pm$0.048 & 0.615$\pm$0.009 & 0.841$\pm$0.007 & \textbf{0.633$\pm$0.067} & 0.662$\pm$0.075 & 0.681 \\
    
    Patches at $20\times$, $10\times$ and $5\times$ & \textbf{0.697$\pm$0.040} & 0.632$\pm$0.035 & 0.856$\pm$0.037 & 0.631$\pm$0.022 & \textbf{0.678$\pm$0.067} & \textbf{0.699} \\
    \bottomrule
  \end{tabular}
  \label{tab:ablation-auc}
\end{table*}

\subsection{Ablation Studies}
To evaluate the impact of hierarchical architecture in solving MIL tasks, we conduct an ablation study to evaluate patches sampled on fixed magnification with pretrained ViT for feature extraction. Table \ref{tab:ablation-c} and table \ref{tab:ablation-auc} shows the C-index and AUC results, respectively, for HiMT models using: 1) patches sampled from single magnification, 2) patches sampled from multi-scale WSIs. The results illustrate that both feature extraction using pre-trained ViT model and hierarchical architecture contribute to the overall c-index. 

Specifically, from Table \ref{tab:ablation-c} and Table \ref{tab:ablation-auc}, the models use patches from all three resolutions have the best overall performance, with an average C-index of $0.673$ and average AUC of $0.699$. In terms of each cancer dataset, although it beat almost all the benchmark methods, the performance of LUAD and BRCA do not show significant improvement with different experimental settings. LUAD stands for lung cancer, which is well-known for the hybrid entities, high genetic instability, high malignancy, and mortality \cite{petersen2011morphological}. Even in the clinic, the diversity causes problems for experts to diagnose. Surprisingly, both the mean C-index and mean AUC of BRCA dataset with $5\times$ magnification achieve the highest score among multi-scale settings. BRCA is a heterogeneous complex of diseases with many subtypes that has distinct biological features \cite{yersal2014biological}. The result might suggest that tissue-level information is more important to HiMT on survival prediction. 
% BRCA from TCGA contains invasive tumors 
% mucinous carcinoma

Noticeably, in the third row of Table \ref{tab:ablation-c} and \ref{tab:ablation-auc} where the model only takes $1000$ $256\times 256$ patches sampled from each $20\times$ WSI as input, HiMT outperforms the MCAT\cite{chen2021multimodal}, which uses over $15,000$ patches per WSI with a similar patch size for feature representation. The result further suggests that pre-trained ViT has out-of-distribution generalization ability and could be used for image feature extraction in a wide variety of research.

\section{Conclusion}

This paper presents a Transformer-based hierarchical architecture for weakly supervised survival prediction on multimodality data.  
Compared with the state-of-the-art methods, our work requires remarkably less computational resources by reducing the number of patches from WSIs without losing performance. Obtaining features from different magnification capture representative information from the microenvironment. 
Our results have quantitatively shown significant improvement over the state-of-the-art methods on survival prediction. 

\textbf{Future work.} The relative position of different types of cells is vital to identifying the cancer stage, which is a critical measurement during diagnosis. Future research could include integrating spatial relations to capture relative position between different cell types, potentially improving further. Besides, the ViT models used for feature extraction are pre-trained on ImageNet-1k, while the pathology image is distinct from the nature image. A ViT model pretrained on an ensemble of cancers might be able to extract more representative patch features.

\section*{Acknowledgment}
This work was partially supported by the NSF CAREER grant IIS-1553687 and Cancer Prevention and Research Institute of Texas (CPRIT) award (RP190107).

The results in this paper are in whole based upon data generated by the TCGA Research Network: \url{https://www.cancer.gov/tcga}.

\bibliographystyle{IEEEtran}
\bibliography{IEEEabrv,refs}

\end{document}